\documentclass{article}

\usepackage[final,nonatbib]{neurips_2022}

\usepackage[utf8]{inputenc} 
\usepackage[T1]{fontenc}    
\usepackage{hyperref}       
\usepackage{url}            
\usepackage{booktabs}       
\usepackage{amsfonts}       
\usepackage{nicefrac}       
\usepackage{microtype}      
\usepackage{graphicx}
\usepackage{xspace}
\usepackage{amsmath}
\usepackage{comment}
\usepackage{xcolor}
\usepackage{mathtools}
\usepackage{setspace}
\usepackage{algorithm}
\usepackage{algorithmicx}
\def\deepwalk{\textsc{DeepWalk}\xspace}
\def\nodevec{\textsc{Node2Vec}\xspace}
\def\line{\textsc{LINE}\xspace}
\def\graphsage{\textsc{GraphSage}\xspace}
\def\gat{\textsc{GAT}\xspace}
\def\gcn{\textsc{GCN}\xspace}
\def\pgnn{\textsc{PGNN}\xspace}
\def\clustergcn{\textsc{ClusterGCN}\xspace}
\def\hgnn{\textsc{H2GCN}\xspace}
\def\rgcn{\textsc{RGCN}\xspace}
\def\netgan{\textsc{NetGAN}\xspace}
\def\graphrnn{\textsc{GraphRNN}\xspace}
\def\deepgmg{\textsc{DeepGMG}\xspace}

\def\molgan{\textsc{MolGAN}\xspace}
\def\graphgen{\textsc{GraphGen}\xspace}
\def\gran{\textsc{GRAN}\xspace}
\def\dyngem{\textsc{DynGem}\xspace}
\def\tnodeembed{\textsc{tNodeEmbed}\xspace}
\def\dynamictriad{\textsc{DynamicTriad}\xspace}
\def\dysat{\textsc{DySAT}\xspace}
\def\evolvegcn{\textsc{EvolveGCN}\xspace}
\def\ige{\textsc{IGE}\xspace}
\def\caw{\textsc{CAW}\xspace}
\def\tgat{\textsc{TGAT}\xspace}
\def\tgn{\textsc{TGN}\xspace}
\def\dyrep{\textsc{DyRep}\xspace}
\def\jodie{\textsc{JODIE}\xspace}
\def\tigecmn{\textsc{TigeCMN}\xspace}
\def\htne{\textsc{HTNE}\xspace}
\def\fitne{\textsc{FiTNE}\xspace}
\def\mdne{\textsc{MDNE}\xspace}
\def\taggen{\textsc{TagGen}\xspace}
\def\dymond{\textsc{DYMOND}\xspace}

\newtheorem{defn}{\textbf{Definition}}
\newtheorem{prob}{\textbf{Problem}}
\newtheorem{ex}{\textbf{Example}}

\newcommand{\mbd}[1]{\mathbf{#1}\xspace}
\newcommand{\z}{\mathbf{z}\xspace}
\newcommand{\Z}{\mathbf{Z}\xspace}

\newcommand{\m}{\mathbf{m}\xspace}
\newcommand{\M}{\mathbf{M}\xspace}

\newcommand{\h}{\mathbf{h}\xspace}
\newcommand{\W}{\mathbf{W}\xspace}
\newcommand{\w}{\mathbf{w}\xspace}

\newcommand{\X}{\mathbf{X}\xspace}
\newcommand{\R}{\mathbf{R}\xspace}

\newcommand{\A}{\mathbf{A}\xspace}
\newcommand{\D}{\mathbf{D}\xspace}
\newcommand{\I}{\mathbf{I}\xspace}
\newcommand{\bH}{\mathbf{H}\xspace}
\newcommand{\x}{\mathbf{x}\xspace}

\newcommand{\s}{\mathbf{s}\xspace}
\newcommand{\e}{\mathbf{e}\xspace}
\newcommand{\ba}{\mathbf{a}\xspace}
\newcommand{\bb}{\mathbf{b}\xspace}

\newcommand{\ve}{\mathbf{v}\xspace}

\DeclareMathOperator*{\argmin}{argmin}

\title{A Survey on Temporal Graph Representation Learning and Generative Modeling}

\author{%
  Shubham Gupta\\
  Department of Computer Science\\
  IIT Delhi\\
  \texttt{shubham.gupta@cse.iitd.ac.in} \\
  \And
  Srikanta Bedathur \\
  Department of Computer Science\\
  IIT Delhi \\
  \texttt{srikanta@cse.iitd.ac.in} \\
}

\begin{document}

\maketitle

\begin{abstract}
Temporal graphs represent the dynamic relationships among entities and occur in many real life application like social networks, e-commerce, communication, road networks, biological systems, and many more. They necessitate research beyond the work related to static graphs in terms of their generative modeling and representation learning. In this survey, we comprehensively review the neural time-dependent graph representation learning and generative modeling approaches proposed in recent times for handling temporal graphs. Finally, we identify the weaknesses of existing approaches and discuss research proposal of our recently published paper \textsc{Tigger} \cite{tigger}.
\end{abstract}


\section{Introduction}
Traditionally static graphs have been the de facto data structures in many real-world settings like social networks, biological networks, computer networks, routing networks, geographical weather networks, interaction networks, co-citation networks, traffic networks, and knowledge graphs \cite{10.5555/2361850,10.5555/1971972,10.5555/1050985}. These graphs are used to represent the relationships between various entities. Major tasks like community detection, graph classification, entity classification, link prediction, and combinatorial optimization are established research areas in this domain. These tasks have applications in recommendation systems \cite{graphrecsys}, anomaly detection\cite{graph_anamoly}, information retrieval using knowledge graphs \cite{Frber2018LinkedDQ}, drug discovery \cite{graphgen} , traffic prediction \cite{traffic_prediction}, molecule fingerprinting \cite{NIPS2015_f9be311e}, protein interface prediction \cite{NIPS2017_f5077839} and combinatorial optimization \cite{Peng2021Jun}. Recently, graph neural networks \cite{kipf2017semisupervised,hamilton2018inductive, gat, gin,pgnn} have been developed to improve the state-of-the-art in these applications. Moreover, much success has been achieved in terms of quality and scalability by the graph generative modeling methods \cite{you2018graphrnn,pgnn,liao2020efficient}. 

However, most of these datasets often have the added dimension of time. The researchers marginalize the temporal dimension to generate a static graph to execute the above tasks. Nonetheless, many tasks like future link prediction\cite{timeawarelinkprediction}, time of future link prediction \cite{knowevolve} and dynamic node classification\cite{jodie} require temporal attributes as well. This has led to recent advancements in temporal graphs in terms of defining and computing temporal properties \cite{Holme_2015}. Moreover, algorithmic problems like travelling salesman problem \cite{MICHAIL20161}, minimum spanning trees\cite{10.1145/2723372.2723717}, core decomposition\cite{7363809}, maximum clique \cite{himmel2017adapting} have been adopted to temporal graphs. Recent research on graphs has focused on dynamic representation learning \cite{tgn,tigecmn,dyrep,tgat} and achieved high fidelity on the downstream tasks. Research in temporal graph generative space\cite{TagGen,DYMOND} is in its early stages and requires focus, especially on scalability. 

Many surveys exist that separately study techniques on graph representation learning \cite{sg1,sg2,sg3,sg4,sg5}, temporal graph representation learning \cite{tg1,tg2,tg3}, and graph generative modeling \cite{gg1,gg2}. But this survey is a first attempt to unify these inter-related areas. It aims to be an initial point for beginners interested in the temporal graph machine learning domain. In this report, we outline the following-
\begin{itemize}
    \item We initiate the discussion with definitions and preliminaries of temporal graphs. 
    \item We then present the summary of graph representation learning methods and the static graph generative methods that are prerequisites to explore similar approaches for temporal graphs.
    \item We discuss in-depth the temporal graph representation approaches proposed in recent literature.
    \item Subsequently, we outline the existing temporal graph generative methods and highlight their weaknesses. 
    \item Finally, we propose the problem formulation of our recently published temporal graph generative model \textsc{Tigger}\cite{tigger}.
\end{itemize}

\section{Definitions and Preliminaries}
This section will formalize the definition of temporal graphs and their various representations. Furthermore, we will explain a temporal graph's node and edge attributes. We will then discuss the various tasks under temporal graph setting and frequent metrics used in literature.
\subsection{Temporal Graph}
Temporal graphs are an effective data structure to represent the evolving topology/relationships between various entities across time dimensions. However, temporal graphs are not only used to describe the phenomenon which simulates the evolving links but also the underlying process which triggers the entity addition or removal from the topology. In addition, they also represent the evolution of entity/edge attributes over time. For example, temporal graph-based modeling can explain the sign-up behavior of a user in a shopping network and the causes of churn apart from their shopping behavior. In this survey, we also use temporal graphs to characterize those dynamic graphs that are not evolving anymore.

A temporal graph is either represented in continuous space or discrete space. In the below subsections, we describe the distinction between the two.
\subsubsection{Continuous Time Temporal Graph}
A continuous-time temporal graph is a stream of dyadic events happening sequentially. 
\begin{equation}
    G = \{(e_1,t_1,x_1),(e_2,t_2,x_2),(e_3,t_3,x_3)....(e_n,t_n,x_n)\}
\end{equation}
Each $e_{i}$ is a temporal event tuple at timestamp $t_i$ with attributes $x_i \in \mathcal{R}^F$. These attributes can be continuous, categorical, both, or none at all. Each event is defined depending on the type of dataset and tasks. For an interaction network like transaction network, shopping network, and communication networks, each $e_i = (u,v)$ is an interaction between a node $u$ and $v$ at a time $t_i$ where $u,v \in V $ and $V$ is a collection of entities/nodes in the network. In a general framework, $|V|$ is a variable across time. Nodes can be added and deleted, which is  represented by the event $e$. An event $e_i =(u,"add")$ is specified as node addition event where a node $u$ with attributes $x_i \in \mathcal{R}^F$ at timestamp $t_i$ has been added to the network. A repeated node addition event is interpreted as a node update event with new attributes if the node is already existing in the network. For example, the role of an assistant professor node can change to associate professor node in a university network.  Similarly, a node deletion event $e_i = (v_i,"delete")$ is also possible. 

In temporal graphs like knowledge graphs, co-citation networks, biological networks, transport network, each edge will have a time-span i.e, an edge $t_{e_i}$ can be added to the network at time $t_1$ and removed from $t_2$ where $ 0 <=t_1, t_2 <= T$. $T$ is the last observed time-stamp in network. In such cases, we will assume that each edge event is either a link addition event ($e_i =(u,v,"add")$) at $t_i$ in the network or a link deletion event ($e_i =(u,v,"delete")$) at $t_j$ in the network.

\begin{figure}[htbp]
    \centering

        \includegraphics[width=2.5in]{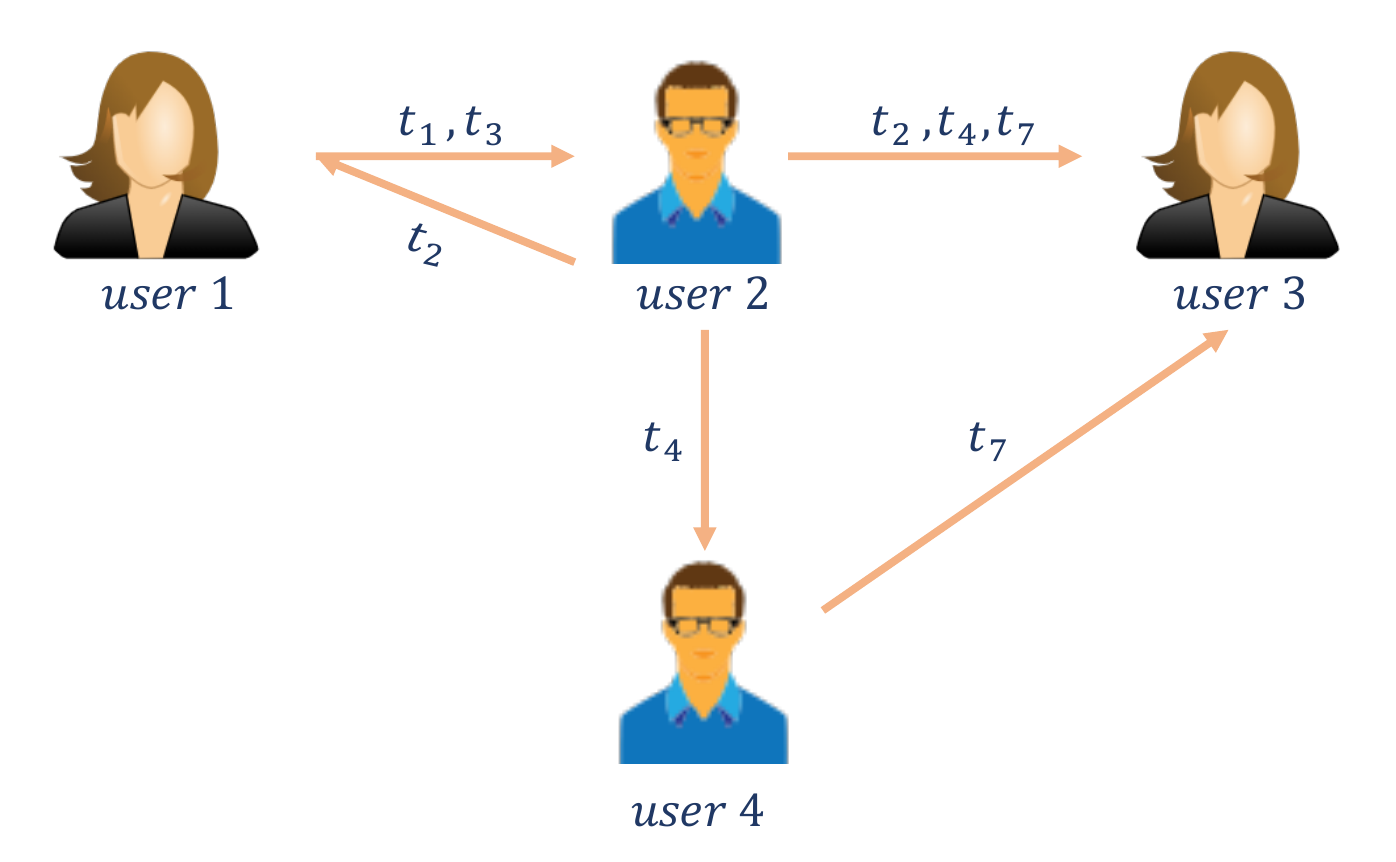}
        \caption{Temporal Interaction Network}
        \label{fig:ctg}
\end{figure}

Figure \ref{fig:ctg} shows an example of one of the interaction networks. Event representation is often designed as per the requirement in the existing literature. For example, \cite{dyrep} add a variable k in the event tuple $(e_i,t_i,k_i)$. Here, $k_i$ is a binary variable signifying whether this event is an association event (the permanent link between two nodes in the network) or a communication event (interaction between two nodes). We encapsulate such intricacies in the variable $e_i$ to simplify the presentation. 


\subsubsection{ Discrete-Time Temporal Graph}
\begin{figure}[ht]
    \centering
        \includegraphics[width=4in]{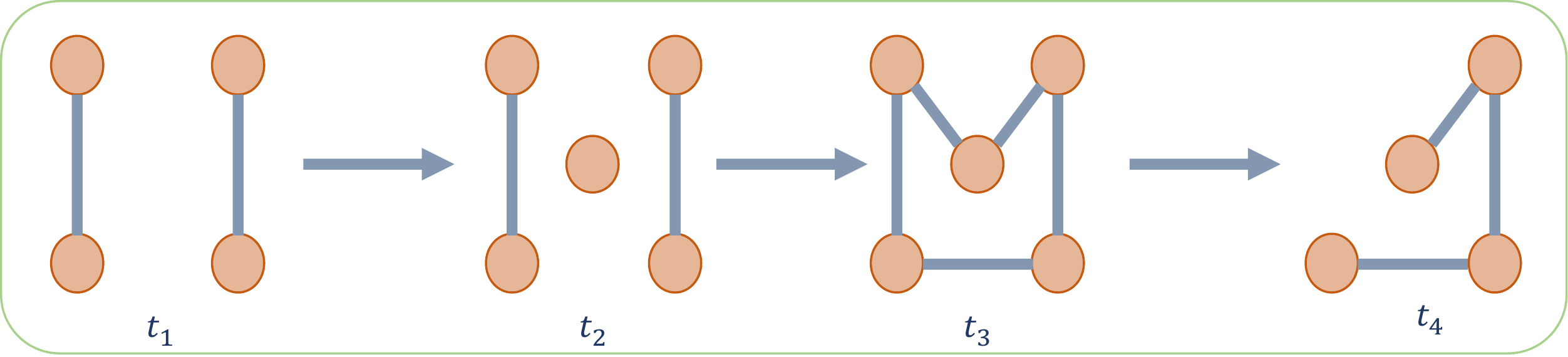}
        \caption{Evolving  Temporal Network}
        \label{fig:dtg}
\end{figure}

Generally, a temporal graph is generated by accumulating evolution across a window of consecutive timestamps to extract the desired information or apply static graph modeling techniques. We divide the time axis in equal lengths and perform aggregation to create a graph along with each such temporal window. Figure \ref{fig:dtg} displays one such aggregated temporal graph. This representation is known as a discrete-time temporal graph. 

A discrete-time temporal graph is defined as follows:
\begin{equation}
    G = \{(G_1,t_1,{X_{1}}^v,{X_{1}}^e),(G_2,t_2,{X_{2}}^v,{X_{2}}^e)....(G_n,t_n,{X_{n}}^v,{X_{n}}^e) \}
\end{equation} 
Here, each $G_i = (V_i,E_i)$ is a static network at time $t_i$ where $V_i$ is a collection of nodes in time-window $t_i$ and $E_i$ is the collection of edges $e=(u,v), u,v \in V_i$. ${X_{i}}^v \in \mathbb{R}^{|V_i| \times F_v}$ is the node feature matrix, where $F_v$ is the dimension of the feature vector for each node. Similarly, ${X_{i}}^e \in \mathbb{R}^{|E_i| \times F_e}$ is edge feature matrix and $F_e$ is the dimension of the edge feature vector. Please note that the meaning of the notation $t_i$ is dataset specific. Essentially, it is a custom representation of the period over which the $G_i$ graph has been observed. For example, $t_i$ can be a month if each snapshot is collected for each month. In many cases, it can simply be the mean of the time-window or last timestamp in the time-window as well. 

\subsection{Attributes}
Attributes are a rich source of information, except for the structure of a graph. For example, a non-attributed co-citation network will only provide information about frequent co-authors and similarities between them, but an attributed network containing roles and titles of every node will also offer a holistic view of the evolving interest of co-authors and help in predicting the next co-author or next research area for co-authors. Attributes, in general, can take a categorical form like gender and location, which are represented as 1-hot vectors. They can also make a continuous form like age. For example, in a Wikipedia network \cite{jodie}, each interaction's attributes are word vector of text edits made in the wiki article. And the class (label) of each user explains whether the user has been banned from editing the page. Sometimes these attributes are present as meta-data in the network. \cite{Frey972} airline network dataset contains the meta-information in the form of the city's latitude, longitude, and population. 

\subsection{Tasks and Evaluating Metrics}

The primary objective in representation learning is to project each node and edge into d-dimension vector space. It achieves it by learning a time-dependent function $f(G,v,t) : \mathcal{G} \times V \times R^+ \rightarrow \mathcal{R}^d$ $ \forall v \in V$ where $V$ is the set of nodes in the temporal network $G$  which has been observed until time T. Generally $t > T$ for future prediction tasks, but time-dependent requirement can be dropped to learn only a node-specific representation \cite{CTDNE}. If the function $f$ allows only an existing node $v \in V$ as an argument, then this setting in literature is generally known as \textbf{transductive representation learning} \cite{hamilton2018inductive}. Often this is desirable not to have this restriction since frequent model training is not possible in many real-life systems, and it is usually required for the trained model to generate representations for unseen nodes $v \not\in V$. This setting is recognized as \textbf{inductive representation learning} \cite{hamilton2018inductive}.  The argument $G$ in the function $f$  encapsulates graph/node/edge attributes and is simplified according to design choices. Suppose we assume that temporal graph in the continuous domain, then $G$ can be approximated as a sequence of event streams or only those events in which another argument $v$ and its neighbors are involved. Neighbors of a node v in a temporal graph do not have a universal definition like static graphs. Most papers typically define the neighbourhood of a node $v$ at given time $t$ as a set of nodes which are $k_d$ hops away in topology from node $v$ and their time of interactions are within $k_t$ of given time $t$. $k_d$, and $k_t$ are application specific parameters. \cite{dyrep,tgat,tgn}. This definition typically selects the recent interactions of node v before time t. \cite{tigecmn,jodie} use the node $v$'s all previous interactions to learn its representation at time t. This is the special case of $k_d = 1$ and $k_t = \infty$.

Every representation learning primarily focuses on learning the node representation and then using these representations to learn the edge embeddings. Edge $e=(u,v,t)$ representation is learnt by learning a function $g(G,u,v,t): \mathcal{G} \times V \times V \times R^+ \rightarrow \mathcal{R}^d$ which aggregates the node representation of its two end nodes and their attributes, including the edge in the consideration. Most often, $g$ is a concatenate/min/max/mean operator. It can also be a neural network-based function to learn the aggregation. We note that t is an argument to the function $g$, allowing the $g$ to learn a time-dependent embedding function. 

Similarly, discrete-time temporal graph snapshots are encoded into low-dimensional representation. Like the edge aggregator function $g$, a graph-level aggregator is learned, taking all nodes in that graph as input.

The most frequent tasks using graph/node/edge representations are graph classification, node classification, and future link prediction. These are known as downstream tasks. These tasks cover many applications in recommendation systems, traffic prediction, anomaly detection, and combinatorial optimization. In recent work, we have also observed additional tasks like event time prediction \cite{dyrep} and clustering \cite{ige}. We will now detail each task and the metrics used to evaluate the model efficiency for these tasks. Please note that we overload the notation $f$ for each task.
\subsubsection{ Node Classification}
Given a temporal graph $G$ and node $v$, a function $f$ is trained to output its label. Formally,
$$
f(G,v,t): \mathcal{G}\times V \times R^+ \rightarrow C 
$$
Where C is a set of categories possible of a node in a temporal graph $G$. Often a node $v$ can be represented as a time-dependent $d$ dimensional vector $\h_v(t)$, $f$ can be approximated to 
$$
f(\h_v(t)): R^d \rightarrow C.  
$$
Node classification can be conducted both in transductive and inductive learning depending upon the argument node to the function $f$. In the transductive setting, the argument node is already seen during training. It is unseen in the case of inductive learning. 
Accuracy, F1, and AUC are frequent metrics for evaluating node classification. The AUC is more favorable since it provides a reliable evaluation even in case high-class imbalance. Anomaly detection is a typical case where class imbalance is observed, and the positive class typically belongs to less than $1\%$ of the population.

\subsubsection{ Future Link Prediction}
Given a temporal graph $G$, two nodes $u$ and $v$, a future timestamp t, a function $f$ is learned which predicts the probability of these two nodes linking at a time $t > T$ where $T$ is the latest timestamp observed in the $G$.
$$
f(G,u,v,t): \mathcal{G} \times V \times V \times R^+ \rightarrow R
$$
Similarly, we can also predict the attributes of this future link. As simplified in the node classification, since the node representations $\h_u(t)$ and $\h_v(t)$ are dependent on $G$ and time $t$, we can write $f$ as follows:
$$
f(\h_u(t),\h_v(t)): R^d \times R^d \rightarrow R
$$
We further observe that in most future link prediction settings, node representations for $t > T$ are approximated as 
$$
\h_v(t) \approx \h_v(T)  \  \forall v \in V
$$
The argument $t$ is often not required in future link prediction functions for such settings since $f$ is simply predicting the probability of link formation in the future. In most methods, $f$ is the cosine similarity function between node embeddings or a neural function that aggregates the information from these embeddings. In some instances, \cite{dyrep}, $f$ is approximated by the temporal point processes \cite{rasmussen2018lecture}, which also allows for predicting the time of link formation as well. Like node classification, we can categorize the future link prediction task into transductive and inductive settings, depending upon whether the node $v$ in consideration is already seen during training. 

Future link prediction is evaluated in 2 significant settings. Researchers typically choose either of these two. In the first set-up, a link prediction task is considered a classification task. The test data consists of an equal number of positive and negative links. Positive links are the actual links present in the future subgraph or test graph. Generally, a test graph is split chronologically from the training graph to evaluate the model performance. Therefore, edges in this test graph or subgraph are considered positive examples. From the sample test graph, an equal number of non-edge node pairs are sampled as negative links. Accuracy, f1, or AUC are used to evaluate the performance of $f$.
In another setting, future link prediction is seen as a ranking problem. For every test node, its most probable future neighbors are ranked and compared with actual future neighbors. In a slightly different framework, the top K possible edges are ranked in a test graph and compared with the ground truth edges. Preferred metrics in these ranking tasks are mean reciprocal rank(MRR), mean average precision(MAP) \cite{10.1145/2939672.2939753}, precision@K, recall@K. 
\subsubsection{ Event Time Prediction}
\cite{dyrep} introduces a rather novel task of prediction of the time of the link in consideration. This task has applications in the recommendation system. Learning which items will be purchased by a user at a particular time t will result in better recommendations, optimized product shipment routing, and a better user experience. Mean absolute error (MAE) is the metric for evaluating this task.

\section{Literature Review}
We first summarize the prevalent static graph representation learning methods. We will later see that temporal graph representation methods are direct extensions of these approaches. 
\subsection{Static Graph Representation Learning Methods}
These methods are divided into two main categories, a) Random walks based methods and b) Graph neural network. There are other categories as well, like factorization based approaches \cite{10.1145/2488388.2488393} \cite{NIPS2001_f106b7f9} and \cite{10.1145/2939672.2939751}. However, these are generally not used due to associated scalability problems and the inability to use available attributes. Moreover, random walk and GNN-based methods are also superior in quality. For this subsection, we assume $G=(V,E)$ as a static graph where $V$ is the node-set and $E=\{(u,v) \mid u,v \in V \}$ is the edge set. $N$ is number of nodes, and $M$ is number of edges. We denote the $1$ hop neighbourhood of node $v$ as $\mathcal{N}_v$ and $\x_v$ as input feature vector for node $v$. Also, the bold small case variable denotes a vector, and the bold large case variable denotes a matrix.  

\subsubsection{Random Walk based Methods}
Node representation is often the reflection of the graph structure, i.e., the more similar the representation, the higher the chances of the corresponding nodes to co-occur in random walks. This intuition provides the unsupervised learning objective to learn the node representation. Building on this, \deepwalk \cite{Perozzi:2014:DOL:2623330.2623732} provided the first random walk-based method. They run the random walks $RW_v$ from each node $v$. Suppose one such k length random walk sequence is $RW_v = \{v_1,v_2 ... v_k\}$. Using the skip-gram objective \cite{word2vec}, they learn the representation $\z_{v} \; \forall v \in V$ by optimizing the following loss objective.
\begin{equation}
\begin{gathered}
L_{RW_v} = -\sum_{v_i \in RW_v}\log P(v_{i-w}...v_{i+w} \mid {v_i}) = -\sum_{v_i \in RW_v} \sum_{v \in v_{i-w}...v_{i+w} \wedge v \neq v_i} \log P(v \mid v_i)\\
p(v \mid u)= \frac{exp({\mbd{z}'}_v^T \mbd{z}_u)}{\sum_{v\prime \in V}exp({\z'}_{v'}^T\z_u)} 
\end{gathered}
\end{equation}
where $\z_v$ is node representation, $\z'_v$ is also the node representation, but it's not used in the downstream tasks. This setup is similar to \cite{word2vec}. $w$ is the window size of window centred at $v_i \in RW_v$. 
$\log p(v\mid u)$ is often re-written  using negative sampling method \cite{NIPS2013_9aa42b31} to avoid the computationally expensive operation in denominator of softmax as follows-
\begin{equation}
\log p(v \mid u) = \log \sigma(\z_v^T\z_u) + \sum_{k=1}^{k=K} \mathbb{E}_{v_n \sim P_n(v)} \log \sigma(-\z_{v_n}^T \z_u)
\label{eq:negative_sampling}
\end{equation}
Where $\sigma$ is the sigmoid function and K is number of negative samples, typically 5, and $P_n(v)$ is a probability distribution over $v \in V$. It is often based on the degree of $v$ and the task. \deepwalk compute these losses for every node in each random walk and update the $\z_v \forall v \in V$ by using gradient descents methods \cite{pmlr-v28-sutskever13}. We note that the next node is selected uniformly from the current node's neighborhood in each random walk. \deepwalk also shows that these learned representations can be utilized in downstream tasks like node classification and missing link predictions. \line is a direct extension of \deepwalk. They modify the \deepwalk loss by restricting co-occurring nodes to be directly connected. Furthermore, they add the following loss as well.
\begin{equation}
\begin{gathered}
L = -\sum_{(u,v) \in E} \log(f(u,v))\\
f(u,v) = \frac{1}{1+\exp(-\z_u^T.\z_v)}
\end{gathered}
\end{equation}
where $E$ is the edge set. This loss forces the neighbouring nodes to have similar representation. 
\nodevec \cite{node2vec}  uses the negative sampling \cite{word2vec} instead of hierarchical softmax to compute the expensive operation in the denominator of softmax. Furthermore, they introduce Breadth-First Search(BFS) and Depth First Search(DFS) biased random walks to learn the node representations. In BFS-based random walks, nodes near in terms of their hop distance are sampled more frequently. This biased sampling leads to learning community structures having similar embeddings for nearby nodes. In DFS-based random walks, nodes which are far are more likely to be sampled. This random walk assigns similar embeddings to nodes having similar roles/structures in the network. \par
These methods directly work with node IDs and do not factor in the node features and associated meta-data. Thus, these approaches are not extendable to unseen nodes in the network since new node IDs are absent during training. Furthermore, these are unsupervised approaches, so embeddings cannot also be learned using available supervision on the nodes/edges. These challenges limit the practical uses of the above methods. 
\subsubsection{Graph Neural Network based Methods}
\cite{kipf2017semisupervised} introduced Graph Convolutional Network (\gcn) to learn the node representation based on graph adjacency matrix and node features. Below equation represents the layer wise message passing in multi-layer Graph Neural Network.
\begin{equation}
    \begin{gathered}
        \bH^{l+1} = \sigma (\tilde{\A}\bH^l\W^l)\\
        \bH^0 = \X
    \end{gathered}
    \label{eq:gcn}
\end{equation}
where $\X$ is a node feature matrix, i.e. each $i^{th}$ row corresponding to the feature vector $\x_i$ of a node $i$, $\bH^l$ is a node representation matrix at $l^{th}$ layer. $\W^l$ is a trainable weight matrix for message passing from layer $l$ to $l+1$. $\tilde{\A}= \D^{-\frac{1}{2}}(\A+\I_N)\D^{-\frac{1}{2}}$ where $\A$ is the adjacency matrix corresponding to the graph $G$. $D$ is a diagonal matrix, where each diagonal element is the degree of the corresponding node in $\A+I_N$ the matrix. $\sigma$ is a non-linear activation like Sigmoid, ReLU etc. Note that $\I_N$ is added to add the self-loops in formulation. This formulation allows the node representation at the next layer to be an aggregation of self features and feature of neighbour nodes. L layer network will cause node representation to be impacted by the L-hops neighbours, which is evident from the formulae. \par
This proposed approach requires supervision, as the input network needs to have a few nodes labeled to learn the $\W^l \; \forall l \in \{0..L-1\}$.
Authors train the $\W$ for each layer by cross-entropy loss after applying softmax over $\h_v^K$ for each labeled node $v$. This approach can not incorporate the unseen nodes as training requires a full adjacency matrix. This requirement also causes scalability issues for large graphs. 

\cite{hamilton2018inductive} identified that equation \ref{eq:gcn} essentially averages the node representation of the target node and its 1-hop neighbor nodes to learn the node representation for each node at the next layer. So, a complete matrix formulation is not needed. Furthermore, the W matrix is also not dependent on node identity. This relaxation motivates the inductive setting as well. \cite{hamilton2018inductive} authors proposed a method \graphsage with the node-level layer-wise message propagation formulation given below.
\begin{equation}
    \h_v^{l+1} = \sigma (\W^l \text{CONCAT}(\h^l_v,\text{AGGREGATE}_l(\{\h^l_u \; \mid \forall u \in \mathcal{N}_v\})))
\end{equation}
where $\h_v^0 = x_v$, $x_v$ is the feature vector for the node $v$. $\text{CONCAT}$ and $\text{AGGREGATE}_l$ are function which are defined as per the requirement. AGGREGATE simply learns the single representation from the set of neighbour node representations. \graphsage uses MEAN, MAX and RNN based aggregator functions. CONCAT is simply the concatenation of two embeddings. Also, note that after each layer propagation, $h_v^l$s are normalized using l-2 norm. 
The above formulation is effective since it allows the weights matrices to adjust importance on the current target node and neighbor nodes representations to learn the subsequent layer representation for the target node. Additionally, this formulation is inductive since node identity or adjacency matrix is not required.  \graphsage utilizes supervised loss on node labels by applying MLP followed by softmax operation to convert the node embedding to a probability vector in node labels space. \graphsage additionally proposes unsupervised loss similar to equation \ref{eq:negative_sampling} where node u and v co-occurs on a short length sampled random walks.\par
In the above formulation, neighbours are treated equal in aggregator functions like Max or Mean Pool. \cite{gat} proposed an attention based aggregator, namely \gat to compute the relative importance of each neighbour. Specifically, the authors proposed the following to compute the node representation of node $v$ at layer $l+1$.
\begin{equation}
    \begin{gathered}
    \h_v^{l+1} = \sigma \left(\sum_{u \in \mathcal{N}_v \cup v}\alpha_{vu}\W\h_u^l\right)\\
    \alpha_{vu} = \frac{\exp(\textbf{a}^T\text{LeakyRELU}(\W\h_v^l \Vert \W\h_u^l))}{\sum_{i \in \mathcal{N}_v \cup v}\exp(\textbf{a}^T\text{LeakyRELU}(\W\h_v^l \Vert \W\h_i^l))}
    \end{gathered}
    \label{eq:gat}
\end{equation}
$\alpha_{vu}$ indicates the importance of message from node $u$ to node $v$. Here, \textbf{a} is a trainable weight matrix. Additionally, \gat  introduces the multi-head attention similar as \cite{NIPS2013_9aa42b31} to utilize the self-attention based learning process. So, the final aggregation function using K attention heads becomes as follows:
\begin{equation}
    \h_v^{l+1} = \|_{i=1}^{i=K} \sigma \left( \sum_{u \in \mathcal{N}_v \cup v}\alpha_{vu}^i\W^i\h_u^l\right)
\end{equation}

Finally, \cite{DBLP:conf/iclr/XuHLJ19} proves that aggregator functions used in \graphsage and \gcn like max-pool and mean-pool are less powerful in graph isomorphism task than sum aggregator by showing that mean-pool and max-pool can produce similar representation for different node multi-sets. They propose the following simpler GNN formulation - 
\begin{equation}
\begin{gathered}
    \h_v^{l+1} = \text{MLP}^{l+1}\left( (1+\epsilon^{l+1})\h_v^l+ \sum_{u \in \mathcal{N}_v}\h_u\right)\\
    \textbf{g} = \|_{l=0}^{l=L} \left(\sum_{v \in G=(V,E)}(\h_v^l)\right)
\end{gathered}
\end{equation}
where g is graph embedding and L is number of layers in GNN and $\epsilon$ is a learnable irrational parameter. They also show that this formulation is as powerful in graph isomorphic test as WL-test if node features are from countable set\cite{leman1968reduction}. \par
The above formulations assign similar embeddings to nodes with similar neighborhoods with similar attributes, even if they are distant in the network. However, embeddings need to account for the node's position in the network in many settings. Applications like routing, which involves the number of hops/distance between nodes for the end objective, are examples of this requirement. Specifically, the node's position in the network should also be a factor in the learned embeddings. \pgnn \cite{pgnn} proposed a concept of anchor nodes to learn position-aware node embeddings. Assuming K anchor nodes as $\{v_1,v_2...v_K\}$ and distances of node $u$ from these nodes respectively as $\{d_{uv_1},d_{uv_2}...d_{uv_K}\}$ then a node $u$ can be represented using a position encoded vector $[d_{uv_1},d_{uv_2}...d_{uv_K}]$ of size K. More anchor nodes can provide better location estimates in different network regions. \pgnn generalizes the concept of an anchor node with an anchor set, which contains a set of nodes. Node $v$'s distance from an anchor set is the minimum of distances from all nodes in the anchor set. We denote $i_{th}$ anchor set as $S_i$. Each $S_i$ contains nodes sampled from $G$. We note that each anchor set can contain a different no of nodes. These K anchor sets create a K size position encoded vector for all nodes. These vectors are used along with original node features to encode each node. However, since each dimension in the position vector is linked with an anchor set, changing the order of the anchor set/position vector should not change the meaning. This constraint requires using a permutation invariant function aggregator in the GNN. So, \pgnn introduces the following formulation for position-aware node representation.
\begin{equation}
    \begin{gathered}
    \h_v^l = \sigma(M_v^l\textbf{w})\\
    M_v^l[i] = \text{MEAN}(\{f(v,u,\z_v^{l-1},\z_u^{l-1}) \;\mid \forall u \in S_i\})\\
    f(v,u,\z_v^{l-1},\z_u^{l-1}) = s(v,u)\text{CONCAT}(\z_v^{l-1},\z_u^{l-1}) \\ 
    s(v,u) = \frac{1}{d_{sp}(v,u)+1}\\
    \z_v^{l-1} = \text{MEAN}(\{M_v^{l-1}[i] \; \mid \; \forall i \in (0...K-1) \}) \\
    \z_v^0 = \x_v
    \end{gathered}
\end{equation}
where $\h^L_v$ is K size position aware representation for node v. \textbf{w} is a trainable parameter. Also, $d_{sp}(v,u)$ is the shortest path between node $v$ and $u$. If $d_{sp}$ is more than a certain threshold than it is assumed to be infinity. This assumption speeds up the all pair shortest path computation.

The above GNNs methods work well in graphs having high homophily. \cite{Zhu2020BeyondHI} defines \textit{homophily} as the ratio of number of edges connecting nodes having the same label to the total number of edges. Many networks like citation networks have high homophily, but networks like dating networks have very low homophily or heterophily \cite{Zhu2020BeyondHI}. They show that state-of-the-art GNN methods work very well on high homophily networks but perform worse than simple MLPs in heterophily networks. Their method \hgnn proposes the following $3$ simple modifications in GNNs to improve their performance in heterophily networks.
\begin{enumerate}
    \item Target node $v$'s embedding (ego embedding) should not be averaged with neighbourhood embeddings to compute the embedding at next layer, as done in \gcn. \graphsage style concatenation is better. This is similar to the skip-connections in deep neural networks, as to increase the depth of the network.
    \begin{equation}
        \h_v^l = \text{COMBINE}(\h_v^{l-1},\text{AGGR}(\{\h_u^{l-1}, \forall u \in \mathcal{N}_v\}))
    \end{equation}
    \item Instead of using 1-hop neighbour's embedding at each layer to compute the ego-embedding, \hgnn proposes to use higher order neighbour as well as follows:-
    \begin{equation}
        \h_v^l = \text{COMBINE}(\h_v^{l-1},\text{AGGR}(\{\h_u^{l-1}, \forall u \in \mathcal{N}_v^1\}),\text{AGGR}(\{\h_u^{l-1}, \forall u \in \mathcal{N}_v^2\})...)
    \end{equation}    
    where $\mathcal{N}^i_v$ denotes the nodes which are at $i$ hops away from node $v$.
    \item Instead of using only the last layer embedding as final embedding for each node, \hgnn proposes combination of representation at all GNN layers.
    \begin{equation}
        \h_v^{\text{final}} = \text{COMBINE}(\h_v^0,\h_v^1 \ldots \h_v^L)
    \end{equation}
    where $\h_v^0=\x_v$ is input feature representation of node $v$.
    
\end{enumerate}

\cite{10.1007/978-3-319-93417-4_38} observed that the proposed GNN architectures apply only to in-homogeneous networks where each node and edge is of a single type. For example, friendship networks. Thus, they proposed a \gcn extension, namely \rgcn, applicable for heterogeneous networks like knowledge graphs where entities and relations can be multiple types. \rgcn suggested that instead of using the same weight matrix for each neighbor during neighborhood aggregation at each layer, use a separate weight matrix for each node type. Since number of unique relations and nodes are typically around millions, they propose either using block-diagonal matrices or decomposing using basis matrices and learning the coefficient of basis matrices for each entity and relations. Finally, \cite{Chiang_2019} proposed a method \clustergcn to learn GNNs on large-scale networks which have millions of nodes and edges. They proposed to first cluster nodes based on any standard clustering approach and then randomly select multiple node-groups from the clusters and create an induced subgraph using these chosen nodes. Now, update the weights of GNN by running gradient descent on this sub-graph. Repeat the process of randomly sampling node groups, creating an induced sub-graph, and training the GNN until convergence. 

\subsection{Deep Generative Models for Static Graphs}
Till now, we have viewed the static graph representation approaches. We now shortly summarize the deep generative models for static graphs. Given a collection of input graphs $\{G_1,G_2...\})$ which are assumed to be sampled from an unknown underlying distribution $p_{\text{data}}(G)$, the goal of generative methods is to learn a probability distribution $p_\theta(G)$ which is highly similar to $p_{\text{data}}(G)$ and produces graphs having highly similar structural properties as input graphs.  Traditional graph generative models assume some prior structural form of the graph like degree distribution, diameter, community structure, or clustering coefficients. Examples include Erdős-Rényi~\cite{Karonski1997} graphs, small-world models~\cite{small_world}, and scale-free graphs~\cite{albert2002statistical}. Prior assumptions about the graph structures can be encoded using these approaches. Still, these approaches do not apply to practical applications like drug discovery, molecular property prediction, and modeling a friendship network. These models cannot automatically learn from the data.
Learning a generative model on graphs is a challenging problem since the number of nodes varies in each graph of the input data. Furthermore, search space and running time complexity to generate edges is often quadratic in $N$ and $M$. Additionally, in naive graph representation, any observed node ordering of a graph has $\frac{1}{N!}$ probability, i.e., a single graph can be represented using $N!$ possible node ordering. So, the learned generative model should be able to navigate this large space, which is not the case with images, text, and other domains.
We initiate the discussion with \netgan \cite{netgan} which learn a generative model from the single input network. Then we compare the methods of \molgan \cite{molgan}, \deepgmg \cite{deepgmg}, \graphrnn \cite{graphrnn}, \graphgen \cite{graphgen} and \gran \cite{gran}. These methods take a collection of graphs as input to learn the generative model.  Among these, \graphgen and \gran currently, are state-of-the-art methods

\netgan samples a collection of random walks of max length $T$ using the sampling approach in \nodevec and train a WGAN \cite{arjovsky2017wasserstein} on this collection. A GAN architecture primarily consists of a generator and a discriminator. The discriminator scores the probability of a given random walk as real. A discriminator is trained by collecting sampled random walks and generated random walks by generator, where its task is to assign high probabilities to real walks and low probabilities to synthetic walks. The discriminator and generator are trained in tandem until the generator generates walks indistinguishable from real walks and confuses the discriminator. The discriminator architecture LSTM based, where each node in a sequence is encoded using a one-hot vector where the size of the vector is $|N|$. Once the LSTM unit processes a sequence, it outputs a logit that provides the sequence scores. Generator architecture is a bit tricky since it involves a stochastic operation of sampling the next node during random walk sequence generation, i.e., $(v_1,v_2 \ldots v_T) \sim \mathcal{G}$. Basically, a vector $\z \sim Normal(\textbf{0},\textbf{1})$ is sampled from the normal distribution. $\z$ is transformed to a memory vector $\m_0=f_z(\z)$ where $f_z$ is a MLP based function. This $\m_0$ is used to initialize the LSTM cell along with $\mathbf{0}$ vector which outputs the next memory state $m_1$ and probability distribution $p_1$ over the next node $v_1$. From this multinomial distribution $p_1$, the next node $v_1$ is sampled, which along with $m_1$ is passed to LSTM cell to output $(m_2,p_2)$. This process repeats until the generator samples the $T$ length node sequence. To enable backpropagation using this method, the next node sampling $v_i$ from multinomial distribution $p_i=(p_i(1),p_i(2)\ldots p_i(N))$ is replaced with Gumbel-softmax trick \cite{jang2016categorical} which is essentially $p_i^{\ast} =  \text{softmax}(\frac{p_i(1)+g_1}{\tau},\frac{p_i(1)+g_1}{\tau} \ldots \frac{p_i(N)+g_N}{\tau})$ where $g_i$s are sampled from Gumbel distribution with 0 mean and 1 scale. Forward pass is computed using $\text{argmax}(p_i^{\ast}$ and backpropagation is computed using continuous $p_i^{\ast}$. $\tau$ is a temperature parameter.   Low $\tau$ works as argmax, and very high $\tau$ works as uniform distribution. Once the WGAN is trained, \netgan samples a collection of random walk sequences from the generator. It creates a synthetic graph by selecting top $M$ edges by frequency counts.  \netgan uses multiple graph properties like max node degree, assortativity, triangle count, power-law exponent, clustering coefficients, and characteristic path length to compare synthetic graph $\tilde{G}$ with original graph $G$. We note that \netgan's node space is the same as the input graph $G$'s node space $V$. Due to this limitation, its usages are limited to applications requiring samples with similar properties as the input graph but with the same node space. Next, we look at the \molgan \cite{molgan} which is similar to \netgan in terms of using GAN. \molgan takes a collection of graphs $\{G_1, G_2 ...\}$ as input instead of a single graph. Further, \molgan is specifically designed for molecular graphs, which will be clear with the design choices.\par
\molgan utilizes the GAN architecture to learn the generator. Generator $g$ takes a noise vector  $\z \sim Normal(\textbf{0},\textbf{1})$ and transform it using MLPs to a probability-based adjacency matrix and node feature matrix. Using Gumbel-softmax, adjacency and node feature matrices are sampled from probability matrices. Finally, a GNN-based discriminator tries to classify actual graphs and generated graphs. As we see, in \molgan generator produces a whole graph, but in \netgan, it was used to produce a random walk. Finally, as \molgan noted that there are available software packages \footnote{http://www.rdkit.org/} to evaluate the generated molecules in terms of desired chemical properties. These scores act as rewards in \molgan to provide additional supervision to the generator. This model's parameters are trained using a deep deterministic reinforcement-learning framework \cite{lillicrap2019continuous}. This method is not scalable since it requires $O(N^2)$ computation and memory. Furthermore, a graph can be represented using $N!$ adjacency matrices, leading to significant training challenges. We now describe methods that solve these problems.\par
\deepgmg \cite{deepgmg} observes graph generation as a sequential process. This process generates one node at a time and decides to connect the new node to existing nodes based on the current graph state and new node state. \deepgmg employs GNNs to model the states. Specifically, assuming an existing graph $G=(V, E)$ having $N$ nodes and $M$ edges, newly added node $v$, it works as follows:
\begin{equation}
    \begin{aligned}
    ((\ve_1,\ve_2 \ldots \ve_N),\ve_G) &= \text{GNN}^L(G)\\
    v_{\text{addnode}} &= \text{MLP}(\ve_G)\\
    v_{\text{addedge}} &= \text{MLP}((\ve_1,\ve_2 \ldots \ve_N),\ve_G)\\
    s_{u} &= \text{MLP}(\ve_u,\ve_v) \forall u \in V \\
    v_{\text{edges}} &= softmax(\textbf{s})
    \end{aligned}
\end{equation}
For each round of new node addition, an L-layer GNN is executed on the existing graph to compute embeddings of nodes and graphs. Using graph embedding, first, a decision is taken to add a new node or not. If yes, then a decision is taken whether to add edges using this node or not. If yes, then using embedding of existing nodes and new nodes, a score is calculated, and subsequently, a probability distribution is computed over the edges of node $v$ with existing nodes. From this distribution, edges are sampled. Moreover, this process repeats till a decision is taken not to add a new node. The embedding of a new node is initialized using its features and graph state. This approach is also computationally expensive, $O(N(M+N))$ since it runs a GNN and softmax operation $O(N)$ over existing nodes for each new node. It has a lower memory requirement since it no longer needs to store the whole adjacency matrix. This process is trained using Maximum likelihood over the training graphs. \molgan and \deepgmg evaluate the generative models' performance by visual inspection and using offline available quality scores provided by chemical software packages. \par
\graphrnn \cite{graphrnn} follows similar graph representation approach as \deepgmg but replaces GNN with Recurrent neural architecture and uses Breadth-First Search(BFS) based graph representation. Furthermore, it introduced a comprehensive evaluation pipeline based on graph structural properties. Its contributions are summarized as follows:
\begin{itemize}
    \item \textbf{Graph Representation:} Unlike previous methods, \graphrnn considers the BFS node ordering of each permutation to reduce the number of possible permutation as many permutations map to a single BFS ordering. Although a graph can still have multiple BFS sequences, permutation space is drastically reduced. This approach has two-fold benefits.
    \begin{enumerate}
        \item Training needs to be performed over possible BFS sequences instead of all possible graph permutations.
        \item A major issue with \deepgmg was that possible edges are computed for each node, with all previous nodes causing $O(N^2)$ computations. But \graphrnn makes the following observation in any BFS sequence, i.e. \textit{whenever a new node $i$ is added in the BFS node sequence $(v_1,v_2 \ldots v_{i-1})$ where it does not make an edge with a node $v_j$ for $j< i$, we can safely say that any node $\{v_k,k\leq j\}$ will not make an edge with $v_i$. } This observation implies that we do not need to consider all previously generated BFS nodes for possible edges with the new node. We can empirically calculate $W$ signifying latest $W$ generated nodes are needed for a possible edge with the new node. This reduces the computation to $O(WN)$.  
    \end{enumerate}
\item \textbf{Hierarchical Recurrent Architecture:} \graphrnn uses two-level RNNs for modelling each BFS sequence. Primary RNN decides the new node and its type. Node type also includes a stop node to signal the completion of the graph generation process. The hidden state of the primary RNN  initializes the hidden state of the secondary RNN, which sequentially processes over the latest $W$ nodes in the BFS order sequence to create possible edges. These 2 RNNs are trained together using the maximum likelihood objective. 
    
    \item \textbf{Metrics:} \graphrnn introduced Maximum Mean Discrepancy (MMD) \cite{10.5555/2188385.2188410} based metrics to compute the quantitative performance of a graph generator. For $I$ input graphs, a node degree distribution is calculated for each graph in the input graph set and generated graph set. MMD is used to compute the distance between these two distributions. \graphrnn shows the MMD distances for degree distributions, clustering coefficient distribution, and four-node size orbit counts.
\end{itemize}
\gran \cite{gran} also follows a somewhat similar procedure as \deepgmg of learning edges for each new node with previously generated nodes during the graph generation process. \gran observes the graph generation process of \deepgmg as creating a lower triangular part of adjacency matrix, i.e., generating 1 row at a time starting from the first row. This process requires $O(N)$ sequential computations. To scale this approach for large graphs $\sim 5$k, instead of generating 1 row at a time, they generate blocks of B rows, i.e., $O(N/B)$ sequential computation. Furthermore, they drop the recurrent architecture to model the sequential steps used in \deepgmg and \graphrnn. This step facilitates parallel training across sequential steps. At each sequential step, the main task is to discover edges between nodes within the new block and edges between existing nodes and new nodes. To do this, \gran creates augmented edges between nodes of the new block. They also create augmented edges between new nodes and all existing nodes. Finally,  they use rows of lower triangular adjacency matrix as features for existing nodes and $\textbf{0}$ for new nodes. They transform these features to low dimensions using a transformation matrix. Finally, \gran runs a $r$ round of GNN updates on this graph to compute each node's embedding. Using these embeddings, they learn the Bernoulli distribution over each augmented edge. These learned distributions are utilized to sample edges. We note that \gran has used MLP over a difference of embeddings of nodes during message passing and computation of attention coefficients. The overall time complexity of \gran is the same as \deepgmg. Since the architecture is parallelizable, \gran can train and generate graphs for large-scale datasets compared to \graphrnn. \par 
Finally, \graphgen introduces a minimum DFS code-based graph sequence representation. Minimum DFS code is the canonical label of a graph capturing structure and the node/edge labels. Canonical labels of two isomorphic graphs are the same. A DFS code sequence is of $M$ length where $M$ is number of edges. Each edge $(u,v)$ having node labels $label(a),label(b)$ and edge label $label(uv)$ in the sequence is represented as $(t_u,t_v, label(u),label(uv),label(v))$ where  $t_u$, $t_v$ is time of discovery of node $u,v$ during DFS traversal. Essentially, DFS codes can be lexicographically ordered. The smallest DFS code among possible DFS codes of a graph is called the minimum DFS Code. Minimum DFS code is an interesting concept, and for thorough details, we refer to \graphgen \cite{graphgen}. This representation drastically reduces the possible permutations of each graph, which speeds up the training process. Using the maximum likelihood objective, an LSTM-based sequence generator is trained over the minimum DFS codes generated from input graphs.

\subsection{Temporal Graph Representation Learning Methods}
We now summarize the temporal graph representation learning based methods. Overall, we can classify these methods into two major categories, \textbf{Snapshot/discrete graph based methods} \cite{dysat,dyngem,dynamictriad,tNodeEmbed,evolvegcn} and \textbf{Continuous time/event-stream based temporal graphs} \cite{MDNE,FiTNE,HTNE,jodie,tigecmn,dyrep,tgat,tgn,caw,ige}.We discuss both of these categories separately. 
\subsubsection{Snapshot/Discrete Graph based Methods}
For the following discussion in this subsection, we assume the following notation- \par
A dynamic graph $G$ is represented as a collection of snapshots $\{G_1,G_2 ... G_T\}$ where each $G_t = (V_t,E_t,\A_t,\X_t)$, $V_t$ is node set at time $t$ and similarly $E_t, \A_t, \X_t$ is edge set, adjacency matrix and node feature matrix at time $t$. $N_t$ and $M_t$ are number of nodes and edges in graph $G_t$. \par
Na\"ive method involves running a static graph embedding approach over each graph snapshot and aligning these embeddings across snapshots using certain heuristics \cite{hamilton-etal-2016-diachronic}. This is a very expensive operation. \dyngem \cite{dyngem} proposes an auto-encoder based approach which initializes weights and node embeddings of $G_{t}$ using $G_{t-1}$. Mainly, an autoencoder network (MLP based) takes two nodes $u,v$ of an edge in $G_t$ represented by their neighbourhood vector $\textbf{s}_u \in \mathcal{R}^{N_t},\textbf{s}_v \in \mathcal{R}^{N_t}$ and computes a $d$ dimensional vector representation $\h_u,\h_v$. A decoder further takes $\h_u,\h_v$ as input and reconstructs the original $\textbf{s}_u,\textbf{s}_v$ as $\hat{\textbf{s}}_u,\hat{\textbf{s}}_v$. Following loss is optimized at each timestamp $t \in [1\ldots T]$ to compute the parameters.
\begin{equation}
\begin{gathered}
    L_t = L^{\text{global}}_t + \beta_1L^{\text{local}}_t + \beta_2L^1_t  + \beta_3L^2_t \\
    L_t^{\text{global}} = \sum_{u,v \in E_t} \| \hat{\textbf{s}}_u - \textbf{s}_u\|^2 + \| \hat{\textbf{s}}_v - \textbf{s}_v\|^2 \\
    L^{\text{local}}_t = \sum_{u,v \in E_t}\|h_u - h_v\|^2
\end{gathered}
\end{equation}
where $L_1,L_2$ are $L1$-norm and $L2$-norm computed over networks weights to reduce the over-fitting. $L_{\text{global}}$ is a autoencoder reconstruction loss at $t$ snapshot and $L^{\text{local}}_t$ is a first order proximity loss to preserve the local structure. We note that parameters of autoencoder for $G_t$ are initialized using parameters of autoencoder for $G_{t-1}$ inducing stability over graph/node embeddings over consecutive snapshots and reducing the training computation assuming graph structures doesn't change drastically at consecutive snapshots. \tnodeembed \cite{tNodeEmbed} is a similar method but instead of using autoencoder they directly learn node embedding matrix $\W_t \in R^{N\times d}$ for each snapshot by optimizing node classification task or edge reconstruction task. They also use the following loss to align the consecutive $\W_t,\W_{t+1}$ as follows:
\begin{equation}
\begin{gathered}
    \R_{t+1} = \argmin_{\R} (\|\W_{t+1}\R - \W_t\| + \lambda \|\R^T\R-\textbf{I} \|)\\
    \W_{t+1} = \W_{t+1}\R_{t+1}
    \end{gathered}
\end{equation}
where first term forces stable consecutive embeddings and second terms requires $\R_{t+1}$ being a rotation matrix. Further, they employ recurrent neural network over node embeddings at each timestamp to connect time-dependent embeddings for each node and to learn its final representation. \dynamictriad \cite{dynamictriad} is a similar method based on regularizing embeddings of consecutive snapshots, but it models an additional phenomenon of \textit{triadic closure process}. Assuming 3 nodes $(u,v,w)$ in a evolving social network where $(u,w),(v,w)$ are connected but $(u,v)$ are not, i.e. $w$ is a common friend of $u$ and $v$. Depending upon the $w's$ social habits, it might introduce $u$ and $v$ or not. This signifies that the higher the number of common nodes between $u$ and $v$, the higher the chances of them being connected at the next snapshot. \dynamictriad models this phenomena by defining a strength of $w$ with $u,v$ at snapshot $t$ as follows:
\begin{equation}
    \s_{uvw}^t= \w_{uw}^t(\h_w^t-\h_u^t) +  \w_{vw}^t(\h_w^t-\h_v^t) 
\end{equation}
where $\s_{uvw}^t \in \mathcal{R}^d$ and $\w_{uw}^t,\w_{vw}^t$ denotes the tie strength of $w$ with $u$ and $v$ respectively at time $t$. Utilizing this, \dynamictriad defines a following probability of $u,v,w$ becoming a close triad at snapshot $t+1$ given they are open triad at snapshot $t$ given $w$ is the common neighbour.
\begin{equation}
    p^t(u,v,w) = \frac{1}{1+\exp{(-\boldsymbol{\theta}\s_{uvw}^t)}}
\end{equation}
Since $u$ and $v$ can have multiple common neighbours, any of the neighbour can connect, $u,v$ thus closing the triads with all the neighbours. But in real world, which neighbour(s) closed the triad is(are) unknown. To accommodate this, they introduce a vector $\boldsymbol{\alpha_{uv}^t}$ of length $B$. $B$ is number of common neighbours of, $u,v$ i.e. length of set $B^t(u,v) = \{w, \;(w,u) \in E_t \wedge (w,v) \in E_t \wedge (u,v) \notin E_t\} $. Formally, $\boldsymbol{\alpha_{uv}^t}=(\alpha_{uvw})_{w \in B^t(u,v) }$ where $\alpha_{uvw}=1$ denotes that $u,v$ will connect at time $t+1$ under the influence of $w$. Finally, they introduce probabilities that $u,v$ will connect at, $t+1$ given that they are not connected and in open triad(s) at the time $t$. 
\begin{equation}
\begin{aligned}
p^t_{+}(u,v) &= \sum_{\boldsymbol{\alpha_{uv}^t}\neq \textbf{0}}\prod_{k\in B^t(u,v)} p^t(u,v,w)^{(\alpha_{uvw})} \times (1-p^t(u,v,w))^{(1-\alpha_{uvw})}\\
p^t_{-}(u,v) &= \prod_{k\in B^t(u,v)}(1-p^t(u,v,w))^{(1-\alpha_{uvw})}
\end{aligned}
\end{equation}
where $p_+,p_-$ denotes the $u,v$ connecting/not connecting at next snapshot $t+1$. Summation over $\boldsymbol{\alpha}$ denotes the iterating over all possible configuration of common neighbour(s) causing the connection between $u,v$. We note that at-least 1 entry in $\boldsymbol{\alpha}$ should non-zero to enable edge creation between $u,v$ at next step. Finally, their loss optimization function is:
\begin{equation}
    \begin{aligned}
    L &= \sum_{t=1}^{t=T} L^t_{triad} +\beta_0 L^t_{ranking} + \beta_1 L^t_{smooth}\\
    L^t_{triad} &= -\sum_{i,j \in S^t_+}p^t_{+}(u,v) -\sum_{i,j \in S^t_-}p^t_{-}(u,v)\\
    L^t_{ranking} &= \sum_{u,v \in E_t,\hat{u},\hat{v} \notin E_t}\w^t_{uv}\max(0,\|\h_u^t - \h_v^t \|^2_2 - \|\h_{\hat{u}}^t - \h_{\hat{v}}^t \|^2_2)\\
    L^t_{smooth} &= \sum_{u \in V_t}(\| \h_u^t - \h_u^{t-1} \|^2_2)
    \end{aligned}
\end{equation}
where $S^t_+$ is set of edges which form at $t+1$ and $S^t_-$ is set of edges not existing at $t+1$. $L^t_{ranking}$ is a ranking loss over edges which preserves the structural information of corresponding snapshot and $L^t_{smooth}$ is embedding stability constraint over embeddings of each node. These methods can't directly utilize the node features and only focuses on the first order proximity of graph structures, ignoring the higher order structures. \evolvegcn \cite{evolvegcn} and \dysat \cite{dysat} utilize the graph neural networks to calculate the node embeddings over each snapshot. Thus, these methods are additionally capable of using node features to model the higher order neighbourhood structure. We note that these features are also dynamic, i.e. can evolve over time. \par
\evolvegcn is a natural temporal extension of \gcn. Equation \ref{eq:gcn} is rewritten as follows:
\begin{equation}
    \begin{gathered}
    \bH_t^{l+1} = \sigma (\tilde{\A}_t\bH^l_t\W^l_t)\\
    \bH_t^{0} = \X
    \end{gathered}
\end{equation}
Where sub-script $t$ denotes the time of the corresponding snapshot, we note that in \evolvegcn formulation, $\W^l_t \; \forall t \in [1\ldots T],l \in [1..L]$ are not learned using GNN but are the output of recurrent external network which incorporates the current as well the past snapshots' information. The parameters of a \gcn at each snapshot are controlled by a recurrent model, while node embeddings at corresponding snapshots are learned using these parameters. Specifically, \evolvegcn proposes two variants of $W^l_t$ calculation.
\begin{enumerate}
    \item This variant treats $\W^l_t$ as hidden state of a RNN cell. Specifically,
    \begin{equation}
        \W^l_t = \text{RNN}(H_t^l,\W^l_{t-1})
    \end{equation}
    where $H_t^l$ is the input to the RNN and $\W^l_{t-1}$ is the hidden state at previous timestamp. This variant is useful if node features are strong and plays important role in end task.
    \item This variant treat $\W^l_{t}$ as output of a RNN cell. Specifically,
    \begin{equation}
        \W^l_t = \text{RNN}(\W^l_{t-1})
    \end{equation}  
    where $\W^l_{t-1}$ is a input which was the output of RNN cell at previous timestamp.
\end{enumerate}
The parameters are trained end-to-end using any loss associated with the node classification/link classification/link prediction tasks. \par
Similar to \evolvegcn, \dysat \cite{dysat} is a gnn  based model, namely \gat. \dysat runs two attention blocks. First, it runs a \gat style GNN over each snapshot separately. This provides the static node embeddings at each snapshot, which captures the structural and attributes based information. Then, \dysat run a temporal based \textit{self-attention} block i.e. for each node at time $t$, this module takes it's all previous embeddings and compute a new embedding using \textit{self-attention} which also incorporate temporal modalities. Specifically,
\begin{itemize}
    \item \textbf{Structural Attention }: For each $G_t \in G=\{G_1\ldots G_T\}$, $h_v$ in equation \ref{eq:gat} is replaced as:
    \begin{equation}
    \begin{gathered}
    \h_{v_t}^{l+1} = \sigma \left(\sum_{u \in \mathcal{N}_v }\alpha_{vu}\W\h_{u_t}^l\right)\\
    \alpha_{vu} = \frac{\exp(a_{vu}\textbf{a}^T\text{LeakyRELU}(\W\h_{v_t}^l \Vert \W\h_{u_t}^l))}{\sum_{i \in \mathcal{N}_v }\exp(a_{vi}\textbf{a}^T\text{LeakyRELU}(\W\h_{v_t}^l \Vert \W\h_{i_t}^l))}
    \end{gathered}
    \end{equation}
    where $a_{vu}$ is a graph input which denotes the weight on the edge $(v,u) \in E_t$. Please note that in \gat message is computed from self node, which is not the case here.
    \item \textbf{Temporal Self-attention:}  We assume the learned node representation using structural attention for each node $v \in G$ as $\bH_v=\{\h_{v_1}^L,\h_{v_2}^L \ldots \h_{v_T}^L\}\; \h_{v_t} \in \mathcal{R}^d$
    and $\Z_v=\{\z_{v_1}^L,\z_{v_2}^L \ldots \z_{v_T}^L\}\; \z_{v_t} \in \mathcal{R}^{d'}$. $\bH_v \in \mathcal{R}^{T \times d}$ is input to the temporal self-attention block and $\Z_v \in \mathcal{R}^{T \times d'}$ is the output.  Also, $\z_{v_t}$ is the final output of node $v$ at snapshot $t$ which will be used for downstream tasks. Following similar design as \cite{attentionisallyouneed}, $\bH_v$ is used as query, key and value, thus the name \textit{self-attention}. 
    
    Specifically, $\W_q,\W_k,\W_v \in \mathcal{R}^{d\times d'}$ are matrices which are used to transform $\bH_v$ to the corresponding query, key and value space. Essentially, query and key are used to compute an attention value for each timestamp $t$ with previous timestamps include $t$. Using these attention values, a final value is computed for the timestamp $t$ by aggregating using attention weights over previous timestamp values, including $t$. Specifically,
    \begin{equation}
    \begin{gathered}
        \Z_v = \beta_v(\bH_v\W_v) ,\quad \beta_v^{ij}=\frac{\alpha_v^{ij}}{\sum_{k=1}^{k=T}\alpha_v^{ik}} \\
        \alpha_v^{ij} = \frac{(\bH_v\W_k)(\bH_v\W_q)^T_{ij}}{\sqrt{d'}} + M_{ij}
    \end{gathered}
    \end{equation}
    where $M_{ij}=0 \; \forall i \leq j$ and  $M_{ij}=-\infty\; \forall i > j$.
\end{itemize}
We note that the entire architecture is trained end to end by running random walks over each snapshot and using similar unsupervised loss as in \deepwalk for nodes co-occurring in the walks.

\subsubsection{Continuous Time Graph/Event-Stream Graph-based Methods} Now, we turn our focus to the temporal embedding method for continuous graphs where each edge between two nodes is an instantaneous event. We will first discuss methods that model evolving network topology \cite{FiTNE, HTNE, MDNE}. Then we discuss methods that additionally model graph attributes as well. These include bi-partite interactions only methods \cite{tigecmn,jodie} and general interaction networks \cite{tgat,caw,dyrep,ige,tgn}.
\htne \cite{HTNE} notes that snapshot-based methods model the temporal network at pre-defined windows, thus ignoring the network/neighborhood formation process. \htne remarks that neighborhood formation for each node is a vital process that excites the neighborhood formation for other nodes. For example, in the case of a co-authorship network, co-authors of a Ph.D. student will be her advisor or colleagues in the same research lab. However, as that student becomes a professor in the future, her students might collaborate with students of her previous colleagues, thus exciting edges between nodes. Furthermore, a neighborhood sequence of co-authors for that Ph.D. student might indicate the evolving research interests which will influence the future co-authors. \par
For the below discussion, we denote a temporal graph as $G= (V,E,X)$ where $V$ is the collection of nodes in the network, $E = \{(u,v,t)\mid u,v \in  V,\; t\in \mathcal{R}^{+}\}$ and $\X \in \mathcal{R}^{N\times F}$ is a feature matrix for nodes where $N=|V|$ and $F$ is the feature dimension.
\htne visualizes the neighbourhood formation as a sequence of events, where each event is a edge formation between source and target node at time $t$. A neighbourhood formation sequence $(\mathcal{N}(v))$ of a node $v$ can be represented as $\{(u_i,t_i)\mid i=1,2....I\}$ where $I$ is number of interactions/edges of node $v$. We observe that a node can interaction multiple times with the same node but at different timestamps. \htne assumes a $\mathcal{N}(v)$ as a event sequence for each node $v$ and models these event sequences using marked temporal point processes(TPP) \cite{rizoiu2017tutorial}. TPP is a great tool to model the sequential event sequence, $\{(e_1,t_1),(e_2,t_2) ... (e_T,t_T\}$ where past events can impact the next event in the sequence. TPPs are mostly characterized by the conditional intensity function $\lambda(t)$. $\lambda(t)$ defines an event arrival rate at time $t$ given the past event history $H(t)$. $\lambda(t)$ number of expected events in a infinitesimal time interval $[t,t+\Delta t]$.
\begin{equation}
\lambda(t \mid H(t))=\lim _{\Delta t \rightarrow 0} \frac{\mathbb{E}\left[N(t+\Delta t) \mid \mathcal{H}_{t}\right]}{\Delta t}
\end{equation}
Particularly, \htne utilizes temporal hawkes point process to model the neighbourhood formation sequence for each node $v$ in the network. Below formulation define the hawkes conditional intensity for edge between node $v$ and node $u$ at time $t$.
\begin{equation}
\begin{gathered}
\tilde{\lambda}_{u \mid v}(t) = \mu_{u,v} \quad +\quad \sum_{\mathclap{w,t_w \in \{(w,t_w) \in N(v) \wedge t_w <t\}}}\alpha_{w,u}\kappa_v(t-t_w)\\
\mu_{u,v} = -\| \h_u - \h_v \|^2 \quad
\kappa_v(t-t_w) = \exp(-\delta_v(t-t_w)) \quad
\alpha_{w,u} = -\gamma_{w,v}\| \h_u - \h_w \|^2\\
\gamma_{w,v} = \frac{\exp(-\| \h_w - \h_v \|^2)}{\sum_{{w' \in \{(w,t_w) \in N(v) \wedge t_w <t\}}}\exp(-\| \h_{w'} - \h_v \|^2)}
\label{eq:htne}
\end{gathered}
\end{equation}
where $\mu_{u,v}$ is the base rate of edge formation event between node $u$ and $v$. $\alpha_{w,u}$ is the importance of node $v'$s historical neighbour $h$ in possible edge creation with node $u$. $\kappa$ is the time decay kernel to reduce the impact of old neighbours in current edge formation. Finally, following loss is optimized to train the node embeddings. 
\begin{equation}
    log L = \sum_{v\in V}\sum_{u,t \in N(v)} \frac{\lambda_{u\mid v}(t)}{\sum_{w\in V}\lambda_{w \mid v}(t)} \quad \lambda_{u\mid v}(t) = \exp(\tilde{\lambda}_{u \mid v}(t))
    \label{eq:loss_htne}
\end{equation}
\htne applies exponential over $\tilde{\lambda}$ to enable positive values for conditional intensity, as negative event rates are not possible. Furthermore, $exp$ enables softmax  loss which can be optimized using negative sampling similar to previously seen approaches to enable faster training. 
\mdne \cite{MDNE} is a similar approach which additionally models the growth of the network i.e. number of edges $e(t)$ at each timestamp $t$ apart from edge formation process. Specifically, given number of nodes $n(t)$ by time t,  the following defines the additional new edges at time $t$:
\begin{equation}
    \Delta e'(t) = r(t)n(t)(\zeta(n(t)-1)^\gamma) \quad r(t)=\frac{\frac{1}{|E|}\sum_{(u,v,t) \in E}\sigma(-\|\h_u - \h_v\|^2)}{t^{\theta}}
\end{equation}
where $\zeta,\gamma,\theta$ are learnable parameters. $r(t)$ encodes the linking rate of each node with every other node in the network.
Additionally, they add neighborhood factors of node $u$ and $v$ while calculating the intensity function for edge formation between node $v$ and $u$ in eq. \ref{eq:htne}. Finally, they add a loss term in eq. \ref{eq:loss_htne} on mean square error of predicted and ground truth new edges at each time t. \fitne \cite{FiTNE} is a random walk-based method that defines a k length temporal walk $W$ over temporal graph $G$ as $\{w_1,w_2\ldots w_k\}\;w_i \in E$ and there is no constraint over times of straight edges. Essentially, \fitne follows a similar approach as \deepwalk,\nodevec. It collects a collection of random walks from temporal graph $G$ and learns node embeddings using a skip-gram-based learning framework by similar unsupervised loss. When running a random walk, a transition probability is defined over consecutive edges as follows:
\begin{equation}
    p(e_{out}\mid e_{in}) = \frac{w(e_{out}\mid e_{in})}{\sum_{e\in I(v_{e_{in}})} w(e \mid e_{in})}
\end{equation}
where  $I(v_{e_{in}})$ is the set of outgoing edges from $e_{in}$. Finally, they have proposed unbiased/time difference-based formulations for $w$. These methods do not utilize the graph attributes and generate static node embeddings. Specifically, learned node embeddings are not a function of time. Thus, we now summarize methods that utilize graph attributes and learn node embeddings as a function of time $t$. \par
\jodie \cite{jodie} and \tigecmn \cite{tigecmn} are bipartite interactions network-based methods with specific focus on user-item interaction paradigm. Likewise, their architecture actively involves design choices for users and items. Consequently, these methods are not applicable in non-bipartite interaction networks. A major difference between bipartite and non-bipartite networks is the no interactions between the same type nodes(user-user, item-item). \jodie remarks that current temporal embedding methods produce a static embedding using the temporal network, as we saw in previous approaches. In recommendation applications, this will ensure similar recommendations even if the user revisits the site after one hr/1 day/1 month, and so on. This problem implies node embeddings should be a function of time, $t$, i.e., node embedding of a user should model the changing intent over time. Additionally, \jodie models the stationary nature of the user's intent. Since users interact with a million items every day, recommendation time should be sublinear in order of number of items. \jodie's architecture models all of these aspects. We note that following notation for a bipartite graph $G=(U,I,E)$ where $U$ is collection of user nodes, $I$ is collection of item nodes and $E$ is collection of interactions where $e \in E$ denotes the observed interaction and is represented as $e=(u,i,t,\x), u \in U, i \in I, t \in \mathcal{R}^{+},\x \in \mathcal{R}^F$. Assuming $\h_v(t)$ as embedding of node $v$ at time $t$, $\h_v(t^-)$ as embedding of node $v$ at time just before $t$ i.e. updated embedding of node $v$ after latest interaction at time $t' < t$ ,$\h_v$ as static embedding of node $v$, \jodie defines following formulation of user and item nodes' embeddings after observing an interaction $(u,i,t,\x)$:
\begin{equation}
    \begin{gathered}
    \h_u(t) = \sigma(\W_1^\text{user}\h_u(t^-)+\W_2^\text{user}\h_i(t^-)+\W_3^\text{user}\x+\W_4^\text{user}\Delta t_u)\\
    \h_i(t) = \sigma(\W_1^\text{item}\h_u(t^-)+\W_2^\text{item}\h_i(t^-)+\W_3^\text{item}\x+\W_4^\text{item}\Delta t_i)\\
    \end{gathered}
\end{equation}
where $\Delta t_u,\Delta t_i$ are time difference of last interaction of $u,i$ with $t$ respectively. $\W^\text{user}$s are modelled using an item RNN and similarly $\W^\text{item}$s using a user RNN. We note that all user's RNNs share the same parameters, and similarly all item's RNNs share the same parameters. Finally, \jodie proposes the following projection operator for calculating the projected node embedding  at $t+\Delta t$ where t is the last interaction time and $\Delta t$ is time spent after last interaction.
\begin{equation}
    \h_v^{projected}(t+\Delta t) = (1+\W_p\Delta t)\h_v(t)
\end{equation}
In order to train the network, \jodie utilizes the future interactions of user nodes i.e. given a last interaction time $t$ with item $i$ for user $u$ and its future interaction time $t+\Delta t$ with item j, can we predict the item embedding $\h_j^{predicted}(t+\Delta t)$ just before $t+\Delta t$, which will same as actual item $j$'s embedding $\h_j \| \h_j((t+\Delta t)^-)$. Predicted item embedding $\h_j^{predicted}(t+\Delta t)$ at time $t+\Delta t$ is calculated given the last interaction of node $u$ at time $t$ with node $i$.
\begin{equation}
    \h_j^{predicted}(t+\Delta t) = \W_1\h_u^{projected}(t+\Delta t) +\W_2\h_i((t+\Delta t)^-) + \W_3\h_u + \W_4\h_i+ b
\end{equation}
The architecture is trained end-to-end using the mean square error between predicted item embedding and actual item embedding. Further, it adds a regularization term over the change in consecutive dynamic embeddings for users and items. \jodie notes that since the architecture directly predicts the future item embedding, using \textit{LSH} techniques \cite{7025604}, the nearest item can be searched in constant time. \tigecmn \cite{tigecmn} utilizes memory networks to store each node's past interactions instead of a single latent vector in order to improve the systems' performance. Specifically, \tigecmn  creates a value memory matrix $\M_u$ for each user $u$ and a value memory matrix $\M_i$ for each item $i$. These memory matrices have K slots of $d$ dimension. Furthermore, there exists a key memory matrix $\M^U$ for all users and another key memory matrix $\M^I$ for all items. Assuming a interaction having feature vector $\x_{u,i}$ between user $u$ and item $i$ at time $t$, memory matrix $M^u$ corresponding to user $u$ is updated as follows:
\begin{equation}
    \begin{gathered}
        \h_{u,i} = g(\W_2(\text{DROPOUT}(\;\W_1[\x_{ui}\| \h_i \| \Delta u]+b)))\\
        s_k = \frac{\M^U(k)\h_{ui}^T}{\|\M^U(k)\|_2\|\h_{ui}\|_2} \quad w_k = \frac{\exp(s_k)}{\sum_{j=1}^{j=K}\exp(s_j)} \quad k = 1,2 \ldots K \\ 
        \e_{u,i} = \sigma (\w_e\h_{u,i}+ \bb_1) \quad \M_u(k) = \M_u(k)\odot (1-w_k\e_{u,i}) \quad k = 1,2\ldots K\\
        \ba_{u,i} = \text{tanh}(\w_a\h_{u,i}) \quad \M_u(k) = \M_u(k) + \ba_{u,i}w_k \quad k=1,2 \ldots K\\
    \end{gathered}
\end{equation}
where $\e_{u,i}$ erases values from $\M_u$ using interaction $(u,i,t,\x_{u,i})$ embedding $\h_{u,i}$. $\ba_{u,i}$ is a add vector which updates the $\M_u$. The same approach is followed to update the value memory matrix $M_i$ for item i.  To compute the user $u$'s embedding at time $t$, its value matrix $M_u \in \mathcal{R}^{K\times d}$ is passed through self attention layer to provide context-aware embedding for each slot $k=1,2 \ldots K$ and finally mean-pooled to provide dynamic embedding $\h'_u$ which is concatenated with static node embedding (computed using transformation on one-hot vector representation) to provide $\h_u$. Finally, this network is trained end to end using cosine similarity between user and item of each interaction in conjunction with samples from negative interactions similar to \graphsage's unsupervised loss. We note that node embeddings learnt in \tigecmn are not function of time, i.e. the embedding corresponding to each node will not change after its last interaction. Another method \ige \cite{ige} uses skip-gram \cite{word2vec} style loss for training embeddings. First, they introduce a induced list $S_u={(u,v_1,t_1),(u,v_2,t_2) \ldots (u,v_L,t_L)}$ containing all interactions by each node $u$. \ige defines context window $W(v)$ of size $C$ for each node $v \in S_u$ using its neighbours in the induced list $S_u$ similar to word2vec method. Finally,\ige uses similar loss as skip-gram using negative sampling with these induced lists. Essentially, \ige observes each induced list as sentence or sequence of words to train their embeddings.

\par
We now summarize the temporal graph representation methods, which utilize both dynamic graph topology and associated attributes to learn the node representation of non-bipartite temporal graphs. Specifically, \dyrep \cite{dyrep} remarks that most temporal graphs exhibit two dynamic processes realizing at different(same) timescale, mainly \textit{association} process and \textit{interaction} process. \textit{Association} process signifies the topological changes in the graph, and \textit{Communication} process implies the interaction/information exchange between connected(maybe disconnected) nodes. \dyrep observes that these two processes are interleaving, i.e., an association event will impact the future communications between nodes, and a communication event between nodes can excite the association event. \dyrep notes that association events have more global impact since they change the topology. In contrast, communications events are local, although indirectly capable of global impact by exciting topological changes in the network. Assuming $G = {V,E}, \;E = \{(u,v,t,k)\},u\in V,v \in V\;t\in [0,T],\;$ and k=0 implies association event and k= 1 implies communication event. We note that permanent topological changes are potentially using k=0. The following formulation shows the node embedding update corresponding to the event $e=(u,v,t,k)$.
\begin{equation}
    \h_v(t) = \sigma(\W_1\h_v(t_p^v)+\W_2\z_u(t^-)+\W_3(t-t_p^v)) \quad \h_v(0) = \x_v
\end{equation}
where $t_p^v$ is the last event time by node $v$ and $\z_u(t^-)$ is the aggregated embedding of node $u$'s neighbourhood just before time $t$. We note that node embedding update formulation comprises three main principles.
\begin{enumerate}
    \item \textbf{Self-Propagation:} A node representation should evolve from its previous representation.
    \item \textbf{Localized embedding propagation:} A event(association/communication) between nodes must be the outcome of neighbourhood of the node through which the information propagated.
    \item \textbf{Exogenous Drive:} Finally, a global process can update the node representation between successive events involving that node.
\end{enumerate}
$\z_u(t^-)$ is calculated using the neighbourhood of node $u$ as:
\begin{equation}
  \z_u(t^-)=max(\{\sigma(q_{ui}(\W_4(\W_5\h_i(t^-)+\bb))), i \in \mathcal{N}(u)\})
\end{equation}
where $\h_i(t^-)$ is the latest node embedding of node $i$ before time $t$. $q_{ui}$ denotes the weight of each structural neighbour of node $u$ which reflect the tendency of node $u$ to communicate more with associated nodes, i.e. $q_{ui}$ is higher for those neighbours to which node $u$ has communicated more frequently. Although, we note that a more weightage should have been provided for neighbours among frequent communication neighbours with the latest exchanges. Anyway, for detailed calculation of, $q_{ui}$ we refer to \textbf{Algorithm 1} of \dyrep. Finally, \dyrep utilizes temporal point processes to model an event $e=(u,v,t,k)$. Formally,
\begin{equation}
    \lambda_k^{u,v}(t) = f_k{\W_6^k(\h_u(t^-)\|\h_v(t^-))} \quad f_k(x) = \phi_k\log(1+\exp(\frac{x}{\phi_k}))
\end{equation}
where $\h_v(t^-)$ signifies the most recent updated node representation of node $v$. $f_k$ is a soft-plus function parameterized for each event type $k$. Finally, \dyrep utilizes  following loss for training models' parameters.
\begin{equation}
    L = \sum_{(u,v,t,k)\in E)} -\log(\lambda_k^{u,v}(t)) + \int_{t=0}^{t=T}\sum_{u \in V}\sum_{v \in V}\lambda^{u,v}_k(t) dt
\end{equation}
We note that the second term in the loss signifies the survival probability for events that do not happen till time $T$. This term is computed by sampling techniques. For more details on the sampling procedure, we refer to Algorithm 2 of \dyrep. Further, this approach can handle unseen nodes, i.e., transductive and inductive settings. We note that \dyrep's methodology requires datasets that contain both evolution events and communication events. We see it as a major limitation since most of the available datasets do not have association events. We now focus on methods that require only interaction events and do not majorly differentiate between association and communication events. \tgat \cite{tgat} is a \textit{self-attention} based method similar to \gat but uses novel functional time encoding technique to encode time in an embedding space. Although previous approaches have projected time into embedding space using separate weight matrices, \tgat utilizes \textit{Bochner's theorem} to propose the following time encoding:
\begin{equation}
    \h_T(t) = \frac{1}{\sqrt{d}}(\cos(\omega_1t),\sin(\omega_1t),\cos(\omega_2t),\sin(\omega_2t) \ldots \cos(\omega_dt),\sin(\omega_dt))
\end{equation}
where $d$ is a hyperparameter and $\{\omega_1 \ldots \omega_d\}$ are learnable parameters. For more details on the derivation of this encoding using \textit{Bochner's theorem}, we refer to \cite{xu2019self}. Finally, \tgat defines neighbourhood of node $u$ at time $t$ as $\mathcal{N}_u(t)=\{(v,t'), (u,v,t')\in E \wedge t' < t\}$. So, \tgat defines the following temporal \textit{GAT} layer $l$ at time $t$-
\begin{equation}
    \begin{gathered}
    \tilde{\h}^{l-1}_v(t) = \h_v^{l-1}(t) \| \h_T(t) \\
    \alpha_u = \frac{\exp((\W_{query}\tilde{\h}^{l-1}_v(t))(\W_{key}\tilde{\h}^{l-1}_u(t))^T)}{\sum_{w\in \mathcal{N}_v(t)}\exp((\W_{query}\tilde{\h}^{l-1}_v(t))(\W_{key}\tilde{\h}^{l-1}_w(t))^T)} \quad u \in \mathcal{N}_v(t)\\
    \h^l_v(t)^{\text{attention}} = \sum_{u\in \mathcal{N}_v(t)} \alpha_u \W_{value}\tilde{\h}^{l-1}_u(t)\\
    \h^l_v(t) = \W^l_2\text{ReLU}(\W^l_1[\h^l_v(t)^{\text{attention}}\|\x_v]+\bb_1^l]) +\bb_2^l \quad \quad \h^0_v(t) = \x_v
    \end{gathered}
\end{equation}
\tgat extends this formulation to multi(k)-head attention by using separate $\W_{query},\W_{key},\W_{value}$ for each attention head as follows:
\begin{equation}
    \h^l_v(t) = \W^l_2\text{ReLU}(\W^l_1[\h^l_v(t)^{\text{attention}_1}\|\h^l_v(t)^{\text{attention}_2}\|\ldots \h^l_v(t)^{\text{attention}_k} \|\x_v]+\bb_1^l]) +\bb_2^l 
\end{equation}
Finally, $\h^L_v(t)$ is the final node representation of node $v$ at time $t$ using $L$ layer \tgat . This model is trained, similar to \graphsage, using link prediction or node classification loss.
\tgat's temporal GNN layer is inductive, i.e., it can predict the embedding of unseen nodes since its parameters are not node dependent. Furthermore, it allows incorporating edge features. Moreover, it can also use evolving node attributes $\x_v(t)$. \par
\tgn \cite{tgn} attempts to unify the ideas proposed in previous approaches and provides a general framework for representation learning in continuous temporal graphs. This framework is inductive and consists of independent exchangeable modules. For example, there is a module for embedding time to a vector. This module is independent of the rest and is replaceable with a domain understanding-based embedding function. They also define a new event type which is node addition/updation $e=(v,t,\x)$ where a node $v$ with attributes $\x$ is added/updated in the temporal graph $G$ at time $t$. \tgn defines the following modules:
\begin{itemize}
    \item \textbf{Memory:} A memory vector $\s_v$ is kept for each node $v$. This memory stores the compressed information about the node $v$ and is updated only during a event involving the node $v$. For a new node, its memory is initialized to $\textbf{0}$.
    \item \textbf{Message Function:} Following messages are computed when a interaction $e=(u,v,t,\x_e)$ occurs.
    \begin{equation}
        \m_u(t) = \text{MLP}(\s_u,\s_v,\Delta t_u ,\x_e) \quad \m_v(t) = \text{MLP}(\s_v,\s_u,\Delta t_v ,\x_e)
    \end{equation}
    where $\x_e$ is corresponding edge attributes and $\Delta t_u$ is the time different between $t$ and last interaction time of node $u$.
    In case, the event is node addition/updation $e=(v,t,\x)$, the following message $\m_v = \text{MLP}(\s_v,\Delta t_v,\x)$ is computed.
    \item \textbf{Message Aggregation:} \tgn processes interactions in a batch instead of one by one to parallelize the process. Given $B$ messages for a node $v$, its aggregated message is calculated at time $t$, given that the batch contains messages from $t^{start}$ to $t$:
    \begin{equation}
        \m^{agg}_v(t) = \text{AGGREGATOR}(\m_v(t_1),\m_v(t_2) \ldots \m_v(t_B)) \quad t^{start} \leq t_1 \leq t_2 \ldots t_B \leq t
    \end{equation}
    \item \textbf{Memory Updater:} After computing messages for each node $v$ involving events in the current batch, memory is updates using $\s_v = \text{MEM}(\m^{agg}_v(t),\s_v)$. \text{MEM} can be any neural based architecture. \tgn notes that recurrent neural architecture is more suitable since current memory depends upon the previous memory.
    \item \textbf{Embedding:} This is the last module whose objective is to compute the embedding of required nodes using their neighbourhood and memory states. Specifically, \tgn proposes the following general formulation to compute the embedding of a node $v$ at time $t$.
    \begin{equation}
        \h_v(t) = \sum_{u\in \mathcal{N}_u(t)}f(\s_v,\s_u,\x_{v,u}(t),\x_v(t),\x_u(t))
    \end{equation}
    where $\x_v(t)$ denotes the latest attributes of node $v$,$\x_{v,u}(t)$ denotes the features of the latest interaction between node $v$ and $u$. \tgn remarks that \tgat layer can be utilized here or any other simpler approximation as well basis the requirement. \tgn also proposes following L layer GNN based formulation which is simple and fast.
    \begin{equation}
    \begin{gathered}
    \tilde{\h}^l(t) = \text{ReLU}(\sum_{u\in \mathcal{N}_u(t)}\W_1[\h_u^{l-1}(t)\| \x_{v,u} \| \text{TIME-ENCODER}(t-t_{v,u})]) \\ \h_v^l(t) = \W_2[\tilde{\h}^l_v(t) \| \h_v^{l-1}(t)] \quad \h_v(t) = \h_v^{L}{t}
        \end{gathered}
    \end{equation}
\end{itemize}
This architecture can be trained using  link-prediction or node classification tasks.  \tgn observes that during training, there is an issue of target variable leak when predicting links in current batch, since memory of nodes will be updated using events of this batch. To encounter this, \tgn creates a message storage which stores the events occurring in last batch corresponding to each node in the graph. Now given the current batch during training, they update the memory of  nodes in this batch using the events from message storage and compute loss in this batch by using link prediction objective. This loss is used to update the model's parameter. Now, message storage is updated for nodes occurring in the current batch by replacing their events with events of the current batch. \par
\caw \cite{caw} critiques that currently proposed methods work well in inductive setting only when there are rich node attributes available. \caw remarks that any temporal representation learning method, even in the absence of node features and identity, when trained on a temporal graph governed by certain dynamic laws, should perform well on an unseen graph governed by the same dynamic laws. For example, if a triadic closure process and feed-forward loop process \footnote{if node a excites node b and node c, then node b and c will also link in the future} govern link formation in the training graph. Then, a trained model on such data should be able to produce a similar performance on an unseen graph governed by these two laws even if node attributes are not available. Current methods rely heavily on attributes,  and thus, an inductive setting cannot model these processes in the absence of node identity/attributes. \caw proposes a method that anonymizes node identity by exploiting temporal random walks. Specifically, a random walk $W = ((w_0,t_0),(w_1,t_1)\ldots (w_k,t_k)),t_0 >t_1>\ldots > t_k,(w_{i-1},w_i)\in E \; \forall i$ backtracks each consecutive step, i.e. each consecutive edge in the walk has a lower timestamp than the current edge. Further, $S_v(t)$ is a collection of $K$ random walks of length $k$ starting from node $v$ at time $t$. $g(w,S_v(t))$ is a $k+1$ size vector where each index $i$ specifies the count of node $w$ in $i^{th}$ position across every $W\in S_v(t)$. So, given a target link $u,v$ at time $t$, $I_{caw}(w,(S_u(t),S_v(t))) = \{g(w,S_u(t)),g(w,S_v(t))\}$ denotes the relative identity of node $w$ wrt to $S_u(t)$ and $S_v(t)$. We note that $w$ must occur at-least in any random walk $W \in S_u(t) \cup S_v(t)$. \caw notes such representation traces the evolution of motifs in random walks without explicitly using node identity and attributes. Finally, each random walk $W\in S_u(t) \cup S_v(t)$ can be represented as:
\begin{equation}
    \begin{aligned}
    W = ((I_{caw}(w_0,(S_u(t),S_v(t))),t_0),&(I_{caw}(w_1,(S_u(t),S_v(t))),t_1)\\\ldots (&I_{caw}(w_k,(S_u(t),S_v(t))),t_k)))
    \end{aligned}
\end{equation}
The following formulation is used to encode the target link involving nodes $u,v$ at time $t$.
\begin{equation}
    \begin{gathered}
    \h_W = \text{RNN}(\{(f_1(I_{caw}(w,(S_u(t),S_v(t)))),f_2(t-t')), \forall (w,t') \in W\})\\
    f_1(I_{caw}(w,(S_u(t),S_v(t)))) = \text{MLP}(g(w,S_u(t))) + \text{MLP}(g(w,S_v(t))) \\
    f_2(\Delta t) = [\cos(\omega_1\Delta t),\sin(\omega_1\Delta t) \ldots \cos(\omega_d\Delta t),\sin(\omega_d\Delta t)]\\
    \h_{S_u(t),S_v(t)} = \frac{1}{2*K}\sum_{W\in S_u(t)\cup S_v(t)} \h_W
    \end{gathered}
\end{equation}
These target link embeddings are usable in future link prediction tasks. Finally, \caw observes that the $f_1/RNN$ layer can additionally incorporate node/edge attributes. We note that the \caw model is not applicable in node classification. 

\subsection{Deep Generative Models for Temporal Graphs}
The current research in deep generative models for temporal graphs is lacking and still in a nascent stage. We briefly overview the non-neural/neural methods below.\par

Erdős Rényi graph model and small-world model are de-facto models for static graphs. However, similar works are not available for temporal graphs. Few works have attempted to formulate the generation process of a network to understand the spreading of diseases via contact using temporal networks. \cite{10.1371/journal.pcbi.1003142} has proposed an approach where they start by generating a static network, and for each link in the network, they generate an active time span and start time. Further, they sample the sequence of contact times from the inter-event time distribution. Finally, they overlay this sequence with the active time span of each link to generate an interaction network. \cite{perra2012a} introduces activity-driven temporal network generative approach. At each timestamp, they start with N disconnected nodes. Each node $i$ b becomes active with probability $p_i$, also defined as an active potential, and connects with randomly chosen m nodes. At the next timestamp, the same process is repeated. \cite{Vestergaard_2014} proposes a memory-based method for both node and link activation. It stores a time delta since the last interaction $\tau_{i}$ for every node i and $\tau_{ij}$ for every link between node i and j. Initially, in an N-size network, all nodes are inactive. A node can become active with probability $bf_{node}(\tau_{i})$ and initiate a link with inactive node j with probability  $g_{node}(\tau_{j})g_{link}(\tau_{ij})$. A link can de-activate with probability $zf_{link}(\tau_{ij})$. Both $b$ and $z$ are control parameters. $f$ and $g$ are memory kernels of a power-law distribution. More recently, \dymond \cite{DYMOND} presented a non-neural, 3-node motif-based approach for the same problem. They assume that each type of motif follows a time-independent exponentially distributed arrival rate and learn the parameters to fit the observed arrival rate.
\taggen \cite{TagGen} models temporal graphs by converting them into equivalent static graphs by merging node-IDs with their timestamps interactions (temporal edge) and connecting only those nodes in the resulting static graph that satisfy a specified temporal neighborhood constraint. They performed random walks on this transformed graph and modified them using heuristics-based local operations to generate many synthetic random walks. They further learn a discriminator to differentiate between actual random walks and synthetic random walks. Finally,  the synthetic random walks classified by a discriminator as real are collected and combined to construct a synthetic static graph. Finally, they convert the sampled static graph to a temporal interaction graph by detaching node identity from time. These approaches suffer from the following limitations: 

\noindent
$\bullet$ {\textbf{Weak Temporal Modeling:} \dymond makes two key assumptions: first, the arrival rate of motifs is exponential; and second, the structural configuration of a motif remains the same throughout the observed time horizon. These assumptions do not hold in practice -- motifs may evolve with time and could arrive at time-dependent rates. This type of modeling leads to poor fidelity of structural and temporal properties of the generated graph. \taggen, on the other hand, does not model the graph evolution rate explicitly. It assumes that the timestamps in the input graph are discrete random variables, prohibiting \taggen from generating new(unseen in the source graph) timestamps. 
More critically, the generated graph duplicates many edges from the source graph -- our experiments found up to $80\%$ edge overlap between the generated and the source graph. While \taggen generates graphs that exhibit high fidelity of graph structural and temporal interaction properties, unfortunately, it achieves them by generating graphs indistinguishable from the source graph due to their poor modeling of interaction times.} 
    
\noindent
$\bullet$ \textbf{Poor Scalability to Large Graphs:}  Both \taggen and \dymond are limited to graphs where the number of nodes is less than $\approx$10000, and the number of unique timestamps is below $\approx$200. However, real graphs are not only of much larger size but also grow with significantly high interaction frequency \cite{Paranjape_2017}. In such scenarios, the critical design choice of \taggen to convert the temporal graph into a static graph fails to scale over long time horizons 
since the number of nodes in the resulting static graph multiplies linearly with the number of timestamps. Further, \taggen also requires the computation of the inverse of a $N' \times N'$ matrix, where $N'$ is the number of nodes in the equivalent static graph to impute node-node similarity. This complexity adds to the quadratic increase in memory consumption and even higher cost of matrix inversion, thus making \taggen unscalable. On the other hand, \dymond has a $O(N^3T)$ complexity, where $N$ is the number of nodes and $T$ is the number of timestamps.%
    
\noindent
$\bullet$ \textbf{Lack of Inductive Modelling:} Inductivity allows the transfer of knowledge to unseen graphs~\cite{hamilton2018inductive}. In the context of generative graph modeling, inductive modeling is required to \textbf{(1)} upscale or downscale the source graph to a generated graph of a different size, and \textbf{(2)} prevent leakage of node-identity from the source graph. Both \taggen and \dymond rely on one-to-one mapping from source graph node IDs to the generated graph and hence are non-inductive.

\section{Proposed Work: Scalable Generative Modeling for Temporal Interaction Graphs}

We now explain the problem statement, for which \textsc{Tigger}\cite{tigger} has proposed a solution.

\subsection{Problem formulation}
\begin{prob}[Temporal Interaction Graph Generator]\hfill
\noindent
\begin{defn}[Temporal Interaction Graph]
\label{def:def_tempInt}
A temporal interaction graph is defined as $\mathcal{G} = (\mathcal{V},\mathcal{E})$ where $\mathcal{V}$ is a set of $N$ nodes and $\mathcal{E}$ is a set of $M$ temporal edges $\{(u,v,t) \mid u,v  \in \mathcal{V},t \in [0, T] \}$. $T$ is the maximum time of interaction. 
\end{defn}

\noindent %
\textbf{Input:} A temporal interaction graph $\mathcal{G}$. \\
\textbf{Output:} Let there be a hidden joint distribution of structural and temporal properties, from which the given $\mathcal{G}$ has been sampled. Our goal is to learn this hidden distribution. Towards that end, we want to learn a generative model $p(\mathcal{G})$ that maximizes the likelihood of generating $\mathcal{G}$. This generative model, in turn, can be used to generate new graphs that come from the same distribution as $\mathcal{G}$, but not $\mathcal{G}$ itself.
\end{prob}

The above problem formulation is motivated by the \textit{one-shot generative modelling} paradigm, i.e., it only requires one temporal graph $\mathcal{G}$ to learn the hidden \textit{joint} distribution of structural and temporal interaction graph properties. Defining the joint distribution of temporal and structural properties is hard. In general, these properties are characterized by inter-interaction time distribution and evolution of static graph properties like degree distribution, power law exponent, number of connected components, the largest connected component, distribution of pair wise shortest distances, closeness centrality etc. Typically, a generative model optimizes over one of these properties under the assumption that the remaining properties are correlated and hence would be implicitly modelled. For example, \dymond uses small structural motifs and \taggen uses random walks over the transformed static graph.

\section{Conclusion}
This survey presents a unified view of graph representation learning, temporal graph representation learning,  graph generative modeling, and temporal graph generative modeling. We first introduced a taxonomy of temporal graphs and their tasks. Further, we explained static graph embedding techniques and, building over them, explained the temporal graph embedding techniques. Finally,  we surveyed graph generative methods and highlighted the lack of similar research in temporal graph generative models. Inspired by this, we introduced the problem statement of our recently published paper \textsc{Tigger}. 
\newpage
\bibliographystyle{plain}
\bibliography{references}

\end{document}